# Shadow of the (Hierarchical) Tree: Reconciling Symbolic and Predictive Components of the Neural Code for Syntax

Elliot Murphy[1,2]

**Affiliations**: 1. Vivian L. Smith Department of Neurosurgery, McGovern Medical School, UTHealth, 1133 John Freeman Blvd, Houston, TX 77030, USA

2. Texas Institute for Restorative Neurotechnologies, UTHealth, 1133 John Freeman Blvd, Houston, TX 77030, USA.

**Correspondence**: elliot.murphy@uth.tmc.edu

**Permanent Address**: 1133 John Freeman Blvd, Houston, TX 77030, USA

**Conflict of Interest**: The author declares no conflict of interest.

**Funding**: This work was supported by the National Institute of Neurological Disorders and Stroke (NS098981).

**Submission Type**: Discussion

**Word Count**: 11,633: 'Introduction' (2026), 'Prospects for ROSE' (2998), 'Deep Learning, Shallow Understanding' (2141), 'A Hybrid Neural Code for Syntax' (3963), 'Conclusion' (505)

**Abstract**: Natural language syntax can serve as a major test for how to integrate two infamously distinct frameworks: symbolic representations and connectionist neural networks. Building on a recent neurocomputational architecture for syntax (ROSE), I discuss the prospects of reconciling the neural code for hierarchical 'vertical' syntax with linear and predictive 'horizontal' processes via a hybrid neurosymbolic model. I argue that the former can be accounted for via the higher levels of ROSE in terms of vertical phrase structure representations, while the latter can explain horizontal forms of linguistic information via the tuning of the lower levels to statistical and perceptual inferences. One prediction of this is that artificial language models will contribute to the cognitive neuroscience of horizontal morphosyntax, but much less so to hierarchically compositional structures. I claim that this perspective helps resolve many current tensions in the literature. Options for integrating these two neural codes are discussed, with particular emphasis on how predictive coding mechanisms can serve as interfaces between symbolic oscillatory phase codes and population codes for the statistics of linearized aspects of syntax. Lastly, I provide a neurosymbolic mathematical model for how to inject symbolic representations into a neural regime encoding lexico-semantic statistical features.

# 1. Introduction

We are currently living in perhaps the most experimentally diverse time in the cognitive neuroscience of language. On an almost daily basis, we read of new neuroimaging reports, but with this buzz there also comes a need for theoretial and conceptual reassessment (Murphy 2025). We increasingly run the risk of deploying sophisticated new tools *sans* hypotheses. As physicists often put it, we risk being labeled the "stamp collectors" of cognitive neuroscience (Bernal 1939: 9).

In this Discussion article, I will assess the prospects for an emerging neural code for natural language syntax ('ROSE'; Murphy 2024). I will use this as a launchpad from which to build more principled relations between syntax, predictive coding, and neural mechanisms driving statistical inferences, alongside other topics becoming increasingly relevant to the use of large language models (LLMs) in the field.

## 1.1. *The Symbolic Species*

The recent ROSE neurocomputational architecture for syntax (Representation, Operation, Structure, Encoding) provides an explicitly *symbolic* account of how humans generate hierarchical phrase structures in natural language (Murphy 2024). This is built on the premise that a core (and deeply puzzling) component of language is its compliance with *structure-dependence* over linear rules: "[W]e ignore the simple computation on linear order of words [adjacency], and reflexively carry out a computation on abstract structure" (Chomsky 2023). This principle is available very early to infants (Perkins & Lidz 2021; Shi et al. 2020). ROSE is in line with the mathematically proven feasibility of unobservable entities, like mental representations, being appropriate objects for scientific study (Piantadosi & Gallistel 2024). It is based on the assumption that the basic data structures of syntax are atomic features (Collins 2024), types of linearly readable mental representations (R) that are coded at the single-unit and ensemble level. Operations (O) transforming these units into manipulable objects accessible to subsequent structure-building levels are coded via high-frequency broadband γ activity. Recursive structural inferences (S) over hierarchical phrase structure, providing instructions for things like headedness in linguistic sets, is derived by forms of low-frequency phase synchronization and phase-amplitude coupling (e.g., low-frequency phase synchronization varies systematically with linguistic structure composition; Brennan & Martin 2020). ROSE also provides forms of causal mechanisms that can coordinate each of these levels: Spike-phase/LFP coupling ensures the operations at O coordinate spike firing of atomic representations at R. Specific types of phase-amplitude coupling over certain left frontotemporal language sites provide a top-down code for reading out the cluster of representations maintained by spike-phase coupling, connecting O to S. Lastly, frontotemporal traveling oscillations transfer the hierarchical phrase structure representations at S to different active workspaces (E). I will refer the reader to the original position paper for further details (Murphy 2024), and will invoke here only the most critical components of ROSE for the current discussion. Given that under many current accounts "architectures for processing symbols *seem* decidedly unbiological" (Piantadosi 2021), ROSE provides a path towards biological plausibility.

ROSE can help us model *vertical* aspects of syntax, like the recursive hierarchical formation of sets of features and phrase structures. This involves explicit symbol manipulation, of the traditional kind (Kolers & Smythe 1984; Murphy & Leivada 2022; Marcus 2001). Yet, further refinement is needed to incorporate relational and horizontal types of information (e.g., linear encoding of morphosyntactic agreement). Another topic that was highlighted for future research in Murphy (2024) was the need for a mathematical formalization of the ROSE architecture. Here, I explore these topics jointly, by arguing that a way to achieve these goals is to integrate associationist and predictive neural mechanisms into the existing ROSE architecture, rather than constructing a separate code for the surface statistics of language.

### 1.2. The Algebraic Mind

In essence, my goal is similar to Marcus's (2001) approach to integrating symbolic and connectionist models of cognition. Marcus argued for a hybrid, neurosymbolic architecture that incorporates meritorious aspects of both symbolic (rule-based) and neural network (connectionist) models (see also the Monte Carlo tree search used in AlphaGo, and the hierarchical conceptual ontology in IBM Watson's grammar rules). This might involve incorporating symbolic systems (e.g., knowledge graphs, logic programming) with neural networks to enhance reasoning and interpretability, potentially reconciling how we can simultaneously model compositionality and gradedness together in the brain. Language provides perhaps the most interesting test case for exploring hybrid models of symbolic and connectionist architectures (Marcus & Murphy 2022), since it exhibits clear symbolic representations (Chomsky 1957, 2013) but also robust effects of statistical inference, e.g., lexical frequency effects (Woolnough et al. 2021, 2024). Many aspects of linguistic semantics, like the parsing of negation (Coopmans et al. 2024), seem to require forms of knowledge that fall outside of connectionist and embodied accounts.

## 1.3. The Phantom Pain

With respect to the cognitive neuroscience of syntax, the distinction between vertical and horizontal syntactic information has been supported via direct cortical recordings, being partitioned into closely neighboring portions of frontotemporal cortex and licensing distinct parsing strategies to isolate (Murphy et al. 2024b; see also Lopopolo et al. 2021). Posterior temporal cortex is increasingly being seen here as the critical code in the core syntax network (Biondo et al. 2024; Fahey et al. 2024; Matchin et al. 2024; Matchin & Hickok 2020; Yu et al. 2024), and is the earliest region to jointly resolve vertical and horiztonal dimensions (Murphy et al. 2024b). Horizontal knowledge is crucial for understanding *surface structure, order and adjacency*, governing how words are placed sequentially in a sentence according to language-specific rules. An integrated measure of processing costs pertaining jointly to these dimensions can explain temporal lobe activity in fMRI (Li & Hale 2019). I will argue that the successes LLMs have recently displayed for modeling lexico-semantic information and dependency relations is likely a factor of their competence for linearized syntactic information, rather than symbolic and compositional representations (Leivada et al. 2023a, 2023b; Linzen & Baroni 2021; but see Marcolli et al. 2025 for a different perspective), serving as a phantom of genuine linguistic understanding rather than the real McCoy. I also note here, as I have done elsewhere, that LLMs cannot in principle provide a 'theory' of linguistic competence (Alers-Valentín & Fong 2024).

It seems possible, however, that LLMs do indeed incorporate some version of vertical syntax (Diego-Simón et al. 2024) – potentially via their attention mechanism – but in a wholly different format from how the human brain represents vertical structure, which explains their failures in basic compositional reasoning (Dentella et al. 2024). Hence, my position will be that artificial models should most reliably be used for exploring horizontal dimensions, and much of the evidence I will leverage below at least indicates some very impoverished and peculiar form of vertical representations in LLMs, but I leave this question open for further debate. Still, I stress again that the evidence presented below indicates that modern language models clearly still struggle with causal reasoning and directed exploration (Binz & Schulz 2023), compositionality (Dentella et al. 2024; Lake & Murphy 2023), analogy and human-like abstractions (Mitchell 2021), amongst other things (for related critical discussion, see Baggio & Murphy 2024). Indeed, with LLMs "complete recovery of syntax might be very difficult computationally" (Marcolli et al. 2025: 13), even if we assume that attention modules can in principle "satisfy the same algebraic structure" as what Marcolli et al. postulate as being necessary for the syntax-semantics interface mapping.

When deep language algorithms are shown to predict semantic comprehension from neural activity, they do so within parietal (angular gyrus and supramarginal gyrus) and medial temporal lobe structures (Caucheteux et al. 2022), not robustly within 'core language' sites. In the case of Caucheteux et al. (2022), this alignment seems to be generated to a large extent by the long-distance dependencies within deep layers of GPT-2, a system that uses unidirectional, forward attention, highly suited to horizontal linguistic information. Predictive processing has also been used to explain artificial model alignment with neural activity (again, across regions like anterior STG and parietal sites) (Schrimpf et al. 2021). Importantly, when measures of surprisal were contrasted with measures of syntactic node counts during sentence comprehension, intracranial electrode sites sensitive to surprisal were found to be generally distinct from electrode sites sensitive to syntactic structure (Nelson et al. 2017; see also Woolnough et al. 2023). Next-word prediction has been shown to be insufficient on its own to account for human syntactic processing (Huang et al. 2024); many other factors conspire here beyond statistics (Fodor & Ferreira 1998).

Even when deep language models converge with neural activation to sentence reading, this seems to mostly occur along a strict division between stimuli-specific visual statistics versus non-stimuli-specific properties. For example, in Caucheteux and King (2022) their visual feature CNN aligns with occipital activity, but their word embeddings and compositional semantics layers seem to activate effectively the entire frontotemporal and lateral parietal surfaces. Yet, lexical access does not require the entire lateral cortical surface (Forseth et al. 2018, 2021; Roos et al. 2023; Yeaton 2024). Indeed, Caucheteux and King (2022) show a clear increase in the size of cortical activity when shifting from visual embeddings, word embeddings and finally contextual word embedding, with the latter yielding the largest widespread activity. But direct cortical recordings in humans suggest the *inverse* pattern for lexical to compositional meaning, with intracranial recordings detailing broader activity patterns for basic lexical processing and a much more constrained spatial *and* temporal profile for compositional semantics and syntax (Murphy et al. 2022b, 2023; Woolnough et al. 2023). This mirrors in some ways the *precision* of the coordinates in conceptual space afforded by syntactic configurations in contrast to single words (i.e., a 'young tall man' fetches a more constrained set of entites than 'man') (Jackendoff 2002). Moreover, these documented effects of composition tend to occur somewhere between 200-500ms (see also Chen et al. 2024), whereas Caucheteux and King (2022) report composition effects beginning at 800ms after word onset, implying that these effects are due to *post-composition* semantics.

Similar criticisms can be applied to other recent work using deep language models. When isolating the temporal scope of language predictions using deep language models, Caucheteux et al. (2023) show that shallow predictions align with activity in speech cortex, but that deep predictions (i.e., multiple words far removed from immediate parsing) lead to activity alignment within superior parietal, superior frontal, inferior frontal, inferior middle temporal and orbitofrontal cortices. Their results for semantic and syntactic predictions reveal widespread sites like motor and visual cortices that do not seem to help constrain our theories of how compositional structure is neurally represented.

What, then, is the alternative? Consider how symbols exhibit a number of advantages over continuous representations (Marcus 2001) which should be of considerable interest to researchers in the cognitive neuroscience of syntax, such as their robustness to noise, reduced bandpass, flexibility, distinguishability, unbounded constructability, compactness, efficacy, and simplification. Adding to transformer language models an inductive bias that encourages the model to explain the data by the means of recursive syntactic compositions improves performance across numerous evaluation metrics (Sartran et al. 2022). Importantly, this is a case of a 'strong' architectural component in these models – weaker forms of these types of syntactic biases lead to less impressive performance (McCoy et al. 2020).

### 1.4. The Present Discussion

Following the order of topics briefly introduced above, I will begin the next section by reviewing current prospects for ROSE, based on recent empirical research that provides further converging or direct evidence for it (Section: 'Prospects for ROSE'). Afterwards, I will discuss how statistical and linear processes can be integrated into this framework, and what the possible role of predictive coding might be across both vertical and horizontal codes for syntax (Section: 'Deep Learning, Shallow Understanding'). Lastly, I will discuss how this case study of syntax can serve as an example for how to integrate components of connectionist and symbolic models (Section: 'A Hybrid Neural Code for Syntax'). Effectively, the goal here is to take a model that is explicitly devised to handle recursive, hierarchical symbolic phrase structures (ROSE) and devise a more direct interface with other components of language that rely more saliently on probabilistic inference. My goal here is to ensure that the symbol-rich ROSE model does not neglect the impressive wealth of evidence for probabilistic reasoning across components of the syntax-phonology interface.

## 2. Prospects for ROSE: Emerging Evidence

The symbolic architecture of ROSE takes as its target the operation MERGE and its associated algorithm for labeling hierarchically formed sets of syntactic objects (Chomsky 2013; Murphy 2025; Pan et al. 2024). This algorithm for what to do with MERGE-generated objects has been claimed to be orchestrated by concerns of computational efficiency, minimal search, and other constraints naturally suited to symbolic systems of syntax. An assumption within theoretical linguistics is that sets need to be categorized in order for semantic/interpretive systems to know that {V, PP} is labeled as a verb phrase and not a prepositional phrase (Murphy 2015a, 2015c, 2025). The 'labeling algorithm' is a means by which linguists can explain, from first principles, grammaticality of particular structures, the distribution of noun phrases, illicit cases of 'movement' (when syntactic constituents are displaced from their originally merged position), and even aspects of language change over centuries (van Gelderen 2023), by distinguishing constituents that are properly labeled from those that are not (i.e., are ungrammatical). Still, it is important to stress that MERGE-based systems can function pefectly well (some argue better; Goto & Ishii 2024) without assuming principles of economy and minimal search for labeling.

Suffice to say, these types of symbolic principles are claimed in Murphy (2024) to be necessary for building a cognitively plausible neurobiology of syntax. Mechanisms like phase-amplitude coupling and spike-phase/LFP coupling are claimed to be strong candidates for implementing processes that afford explicit syntactic representations and grammatically-derived semantic intensions, given the well-documented 'managerial' and 'orchestrating' role for top-down oscillatory dynamics (Kazanina & Tavano 2023a; Mendoza-Halliday et al. 2024). In other domains, while self-supervised learning of auditory chunks in deep learning models can lead to the representation of sequences, probabilistic chunks and some elements of algebraic structures (Orhan et al. 2024), the format of these algebraic expressions is notably distinct from the types of syntax-specific knowledge that is the target of ROSE. But what is this knowledge?

## 2.1. Foundations in Syntax

I have previously explored the formal properties of syntax that are typically considered the target of theoretical and psycholinguistic inquiry, centered around a MERGE-based syntax (see Murphy 2024, Section 7.1) and the Universal Generative Faculty (Hauser & Watumull 2017). MERGE can be formalized as follows (WS = workspace; P/Q = workspace objects; X = additional elements):

WS = {P,Q, …}

MERGE(P,Q,WS) = WS' = {{P,Q},$X_1$, …,$X_n$}

This essentially takes two objects inside a workspace and maps them to a new workspace in which they form a set (for further details, see Murphy 2025). MERGE is thus non-associative, commutative and generates nonplanar trees. A planar embedding only arises during externalization via sensorimotor compression and linearization of these hierarchical objects (Kayne 1994). MERGE can derive the assignment of semantic roles ('$\theta$-roles') and force-/discourse-related functions (clauses). For these objects to be semantically interpreted, MERGE-generated objects need to be 'labeled', identifying a set {A, N} as a noun phrase and not an adjective phrase, to deliver appropriate semantic roles. This labeling algorithm is another clear example of symbolic knowledge, often thought to emerge purely via the guidance of economy principles in a graph-theoretic tree-based search procedure (but see Brody 1998; Goto & Ishii 2024 for alternative accounts) and forming part of the core language system. In Murphy (2015a), I proposed the Labeling Hypothesis ("The operation Label constitutes the evolutionary novelty which distinguishes the human cognome from non-human cognomes"), and Hornstein (2009, 2024) has defended this comprehensively (see also Murphy 2019). When reviewing 'the Merge hypothesis' within the minimalist program, Hornstein (2024) comes to the conclusion that in fact "labeling is the key linguistically bespoke operation", since labeling is ultimately the operation critical to closing MERGE and therefore delivers a recursive system of unbounded hierarchy. For Hornstein (2024), "[r]ecursion is a consequence of closure afforded by labels". Labels are useful for showing how dominance relations are "exploited" by grammar (Hornstein 2009: 46). In this sense, modern linguistics has exposed "a new aspect of the world" (Mukherji 2010: 27) – the labeling of MERGE-generated objects. Labeling provides rigid categorization (dominance by type) and the ability to saturate dyadic concepts, going beyond basic monadic concepts (Pietroski 2018). It is also labeling that permits compositional semantics, since it produces a representational result that cannot be reduced to its parts (Hornstein & Pietroski 2009). Labeling breaks the bounds of semantic monotonicity by projecting new categories upon the merging of two syntactic objects, giving us different kinds of 'things' to think about. It allows us to form *inductive definitions*, which for Hornstein (2024) is the major contribution of labeling. As others have summarized, "labels belong to the core part of the grammar that […] cannot be relegated to the interface" (Cecchetto & Donati 2015: 31).

Under ROSE, theories of labeling and MERGE point to a finite series of computational operations available to the human nervous system. Importantly, labeling and MERGE deliver vertical, hierarchical syntactic knowledge, while other operations like Agree (or morphosyntactic effects yielding long-distance agreement relations) are thought to emerge due to non-linguistic sensorimotor 'interface' constraints, and deliver horizontal syntactic information (Murphy & Shim 2020; Narita 2014). It is within this latter domain that statistical learning has been shown to be helpful in resolving agreement relations (Diego-Simón et al. 2024), albeit only the more simple forms of dependencies (and not multiply-embedded levels of long-distance relations) (Linzen & Baroni 2021). The kinds of horizontal 'spreading activation' and priming (Collins & Loftus 1975) often found in the psycholinguistics literature also crop up in deep language models (Sinclair et al. 2022), again suggesting some possible representational overlap for horizontal types of semantic search. Importantly, Sinclair et al. (2022) report strong effects of structural priming in artificial language models (recoverable from sequential information), but much less clear evidence (and some counter-evidence) for the representation of hierarchical syntactic information driven by structural complexity.

A far off goal here would be to ensure that "facts about the brain would select among theories of the mind that might be empirically indistinguishable in other terms" (Chomsky 1986: 40). Currently, it is difficult to isolate distinct neural mechanisms that are sympathetic to different computational accounts of MERGE. For instance, set theory (Chomsky 2013), Hopf algebra (Marcolli et al. 2025), mereological models (Adger Forthcoming) and category theory (Coopmans et al. 2023a; Phillips 2020) all seem amenable to invoking phase-locking and synchronization (respectively: set-formation or 'FormSet' within minimalist syntax; Hopf multiplication; part-whole integrations; category-theoretic morphisms) and cross-frequency coupling (respectively: endocentric labeling; iterative Hopf algebra or co-multiplication; recursively nested parts; monoidal categories forming via tensor products). Consider also systems rooted in combinatory principles like combinatory logic (Piantadosi 2021) that minimally encode functions (i.e., function application over binary branching trees) without the need to stipulate variables (deviating here from Marcus 2001). These systems could also be amenable to grounding some very primitive components of language like 'FormSet', though likely not much else of MERGE-based syntax. ROSE has initially been built from set theory, but it will be of interest to explore the strengths and weaknesses of modelling neural dynamics of ROSE via other formal accounts moving forward. For instance, many tensor models of syntax ignore the 'phasal' nature of syntactic derivations (Marcolli et al. 2025: 128), and they exhibit an inability to capture the proper similarity structures when composing concepts (for more limitations, see Martin & Doumas 2020), and they sideline other processes related to composition, such as those that lead to semantic blending, complex polysemy, copredication, or context-dependent meaning.

Evaluating the benefits of various formal models continues to be an essential first step in navigating this space. Certainly, as a comparison between basic set theory and other more recent formalizations of MERGE can show, *providing a more restrictive and clear mathematical model of syntax will inevitably aid the search for the neural code for language*. Yet, cognitive neuroscience research typically only offers rather vague characteristics for syntactic composition (e.g., the usual choice of 'sentence > word list' analyses) (Murphy & Woolnough 2024).

## 2.2. ROSE as a Plausible Model for Syntax

Turning to newly-emerging empirical support for ROSE, consider Weissbart and Martin (2024). The authors examine phase-amplitude coupling in MEG during speech perception and "challenge a strict separation of linguistic structure and statistics in the brain, with both aiding neural signal reconstruction". Specifically, they argue that an integration process between statistics and structure might be achieved via cross-frequency coupling. They demonstrate a fronto-temporal wave propagating from posterior to anterior MEG sensors in $\delta$ and in the opposite direction for $\theta$, which under ROSE encodes structural information in a workspace. $\delta - \beta$ and $\theta - \gamma$ phase-amplitude coupling in posterior temporal regions and the temporo-parietal junction was connected to syntactic node closures, while $\delta - \beta$ phase-amplitude coupling was also linked to ongoing syntactic complexity (node depth) in these regions in addition to inferior frontal cortex, supporting parts of the putative neural code and cortical localization assumed in Murphy (2024). Meanwhile, syntactic and statistical features were jointly represented in inferior frontal and anterior temporal cortices, pointing to likely regions of a hybrid interface between symbolic neural outputs and perceptual inferences. Interestingly, it was only in the $\delta$ band that rule-based syntactic features were represented slightly greater than statistical features, suggesting that statistical properties of the input were being used to cue symbolic categorial inferences. The authors argue that cross-frequency coupling can help provide "a framework wherein the syntactic structure and statistical cues are jointly processed during comprehension". This claim fully aligns with my approach here.

Phrase-rate neural tracking (measured via evoked power and inter-trial phase coherence) has been found to be sensitive to the *type* of syntactic structure being parsed in speech (the position of the head in two-word phrases) in scalp EEG (Zhao et al. 2024b). This again points to the possibility that the integration of symbolic and probabilistic knowledge interfaces at the level of oscillatory phase codes.

In related work, taking the narrow computational components of MERGE, labeling, and other sub-components of syntax seriously (rather that treating syntax as a monolithic entity), Zhao et al. (2024a) focused on the neural representation of headedness (labeling decisions), and demonstrate sensitivity via inter-trial phase coherence in two components, straddling $\delta/\theta$ (~1-6 Hz) and the $\alpha/\beta$ bands (~10-17 Hz) in scalp EEG to the representation of the head of phrase. This is achieved by a reactivation specifically of the head and not other phrasal elements during the parsing of subsequently connected adverbs to the verb. In principle, this type of research focusing explicitly on headedness could at some point be related to current mysteries in theoretical linguistics, e.g., how and why certain syntactic heads are initially used to drive structure building (which heads get entered into the workspace first).

Notice here that multiple of these studies show effects in the $\delta$ and $\beta$ bands not just for node closure, but also for other core elements of syntax like headedness and tree depth, in line with ROSE.

In other recent work, Ten Oever et al. (2024) show that a phase code of the kind evident in processes like attention and navigation also orchestrates phonological and lexical processing in language, "such that highly probable events appear to be coded at lower excitability phases". Ambiguous words presented at different phases (via neural entrainment or extracting the phase from MEG) are interpreted as one or another word depending on the phase of presentation. Ten Oever et al. (2024) note that "oscillations as a memory phase code is rarely considered", and their results neatly support core claims underlying ROSE, such as the notion of a memory phase code for lexicality (Murphy 2020b: 172–173). As such, this provides additional empirical support for the precise formulation of the R and O levels (Murphy 2020b, 2024).

Dekydtspotter et al. (2024) directly test ROSE through examining multi-cycle *wh*-dependencies in L1 and L2 French speakers, and provide support via scalp EEG specifically for the O component. They then innovate on this component and show how O can also accommodate properties of relational grammatical processing, much in line with how I am aiming here to integrate non-symbolic properties.

Single cells in the hippocampus have recently been found to be sensitive both to individuals and pronouns referring to those individuals (Dijksterhuis et al. 2024), supporting the current formulation of R (Murphy 2024) under which units encode relevant sets of conceptual features that conspire to form the basis of lexical items. Other single-unit research has exposed single cells within core frontotemporal language cortices that respond in an amodal manner to basic linguistic input (Lakretz et al. 2024). Middle temporal cortex showed a preference for questions over declarative sentences, potentially indicating some sensitivity to syntactic features that are called upon uniquely by syntactic movement operations (Adger & Svenonius 2011), or at least some sensitivity to the statistical differences in configuration type, and also suggesting a critical role for this region in core syntax. Importantly, this was the only example of a strong single-unit preference for a higher-order linguistic feature (this one example can potentially be explainable via the presence of a question-specific feature calculus, i.e. [+Q], [+*wh*]), in contrast to widespread single unit sensitivity for statistical features, suggesting (as ROSE would predict) that units encode at the R level atomic features that conspire into bundles of objects accessible to structure-building mechanisms at S and E. Lakretz et al. (2024) summarize: "[I]n contrast to orthographic and phonological features, we have not identified, in the present dataset, any single cell selective to specific values of higher-level linguistic features, such as syntactic ones" (e.g., grammatical number, gender, tense, syntactic embedding, movement, transitivity). Reporting this effect (or lack thereof) in the context of a dataset of 21 awake neurosurgical parients provides strong empirical support for ROSE. In Murphy (2024), one central prediction was the following: "I am predicting here that single-unit recordings will *not* be able to detect effects of accessing a single linguistic feature and then utilizing it in multiple (underspecified) ways across linguistic sub-systems (e.g., syntax-form mappings)". Specifically, the R level is composed of "ensembles of temporarily cooperating neurons" that encode "bundles of features" (e.g., $v^*$, $T_{past}$, d, $v_{unerg}$, A, $C_{rel}$), specifically atomic "lexico-semantic features" (Murphy 2024; see Collins 2024) – pointing to a distributed code for syntactic features of the kind Lakretz et al. (2024) suspect is at play. In summary, Lakretz et al. (2024) reported a lack of selectivity to any single syntactic feature at the single unit level, potentially indicating a role for ensembles of cells in representing feature bundles in higher-order syntax-semantics.

Turning to other concerns, some authors have argued that phase-amplitude coupling is only suited to encoding sequential relations (Coopmans et al. 2023b). Yet, as argued in Kazanina and Tavano (2023b) and Murphy (2020b, 2024), phase-amplitude coupling is far from being limited to this. Exploring this further, and investigating the negotiation between perceptual inferences and grammatical processes, will require the exploitation of postulated symbolic components of lexico-syntactic information (Pietroski 2018; Pustejovsky & Batiukova 2019), moving beyond parameters that pertain more directly to surface statistical features, such as lexical selectivity, phonological density, lexical frequency, semantic familiarity, orthographic neighborhood, word length, and so forth (Forseth et al. 2021; Woolnough et al. 2021).

As mentioned, one of the predictions emerging from ROSE is that "single-unit recordings will *not* be able to detect effects of accessing a single linguistic feature and then utilizing it in multiple (underspecified) ways across linguistic sub-systems (e.g., syntax-form mappings)" (Murphy 2024). Rather, case-by-case syntax-form mappings will be implemented via spike-phase/LFP coupling, and single-unit responses will exhibit sensitivity to more general activity patterns that are triggered across multiple contexts of use for that given representation. Synaptic weights here would rapidly change to capture statistical dependencies encountered in data, but those weights would still be under the purview of spike-phase coupling which, via ROSE, is functionally coupled with low-frequency phase codes indexing structural/hierarchical inferences. Different spiking shapes (regular-spiking, fast-spiking, positive-spiking, etc.) could then be responsible for coding various statistical features of localized representations, a proposal building on single-unit recordings within STG during speech perception that found that speech-evoked responses were represented by all waveform shapes (Leonard et al. 2024). Though further research at the single-unit level is needed to explore the precise mapping between statistical and structural information, for now it seems clear that high frequency γ power in pSTG can encode both statistical and structural properties of musical (melodic) and syntactic structures (McCarty et al. 2023). It is not unreasonable to assume that γ could therefore serve to index the integration of these feedforward and feedback transmissions (at the O level in ROSE, represented via sensorimotor γ transformations), being in a direct line of communication with both 'lower' spiking activity and 'higher', global low-frequency phase codes.

These ideas also relate to some recent results that have resolved in MEG the temporal evolution of linguistic information, from acoustics to compositional syntax. The Hierarchical Dynamic Coding (HDC) framework of Gwilliams et al. (2024) suggests that "each linguistic feature is dynamically represented in the brain to simultaneously represent successive events", without interference across levels. This framework can elegantly account for the temporal progression from minimal to complex linguistic units, and the model I will develop here, grounded in ROSE, can serve as a potential scaffold to assign these differing levels of linguistic structure. That is to say, the temporal unfolding of the activation of linguistic features documented in MEG by Gwilliams et al. (2024) needs to be ultimately grounded in neurobiological mechanisms. For instance, HDC currently only assumes distinct "neural configurations" for each level of linguistic structure (clusters of active cells; Gwilliams et al. 2024, Fig. 1), which could be grounded not just in temporal dynamics but in the *spatiotemporal* architecture of ROSE. By using GloVe vectors and English Lexicon Project-derived statistics for lower-level features, and rule-based symbolic structure for syntactic information, Gwilliams et al. (2024) detail rapid, short windows for acoustic and lexical frequency effects and longer durations for features like syntactic state. Their results provide rather forceful evidence for the importance of symbolic syntactic structures in the activity of neuronal populations.

This takes care of recent developments. But what about more looming questions that concern symbolic-connectionist integration?

## 3. Deep Learning, Shallow Understanding

There are a number of ways we could embed probabilistic activation functions and frequentist, associationist mechanisms (LeCun et al. 2015; Pitkow & Angelaki 2017) inside ROSE, such that symbolic representations become the *targets* of perceptual inference. There is a range of evidence that higher-order symbolic representations, from sentential and discourse representations (Martin 2018) to morphology (Gwilliams et al. 2018), interact with statistical learning mechanisms and predictive coding-type mechanisms to bias perception. Just as how models in the cognitive (neuro)science of language should always consider constraints on computation (van Rooij et al. 2019), they should also wrestle with limits placed on the neurobiology of perceptual inference (Friston 2010; Murphy 2020a; Tavano et al. 2022; Walsh et al. 2020).

Recently, an attempt was made to reconcile certain formal computational components of rationalist vs. empiricist, and symbolic vs. probabilistic models of natural language syntax, via applied constraints from complexity theory onto a MERGE-based syntax (Murphy et al. 2024a). Here, I will attempt a similar type of reconciliation, but instead of focusing on the computational Marrian level I will keep to the implementational level of neurobiology. Tying results from the 'algorithmic' psycholinguistic level of real-time parsing will also be needed in the future (see Hale et al. 2022; Kazanina & Tavano 2023a). In Murphy (2024), focus was placed on basic syntax and the syntax-semantics interface, due to the importance of this in generating compositional semantics (Balari & Lorenzo 2013; Murphy 2021, 2023; Pietroski 2018). Here, I will turn my attention to the syntax-phonology interface.

Crucially, what I have been referring to as horizontal aspects of syntax (dependency tracking and linearly encoded elements of morphosyntax) are plainly not *entirely* statistically driven. Many horizontal dimensions of syntax interface with, and are directed by, vertical phrase structure. We have to wrestle with the *empirical* issue that representations of modern AI systems and artificial language models often tend to converge with those of the brain (Caucheteux et al. 2022) (though not always in the brain areas typically expected), alongside the *conceptual-theoretical* status of symbolic models of syntax as being the most comprehensively explanatory and elegant theories of human cognition (Everaert et al. 2015).

### 3.1. The Syntax-Phonology Interface

This 'vertical-leading' model accounts for how syntactic instructions carve the path that both semantics and certain components of phonology must follow (e.g., intonation, stress, prosody) (Frota & Vigário 2018). Syntax often informs phonology through prosodic structures (Langus et al. 2012), which are hierarchical domains such as intonation phrases (IPs), phonological phrases (φ), and prosodic words (ω). These prosodic domains are influenced by syntactic structures but do not align with them in a one-to-one manner (Samuels 2011). Intonational boundaries are often determined by syntactic structure (e.g., "Saul, who likes Mike, is a lawyer"). Syntactic constituents are mapped to prosodic units in various ways detailed by linguistic theory. For our purposes, I will merely highlight the likely neurobiological locus of these interacting levels of linguistic structure. Importantly, the vertical-leading model I am discussing here is fully compliant with typical assumptions in syntax-phonology interface research, according to which certain types of syntactic information is accessible to phonology, but syntax is strictly phonology-free (Frota & Vigário 2018). This forms the core of Pullum and Zwicky's (1988) *Principle of Phonology-Free Syntax*, and is related to the minimalist model of syntax of Spell-Out by phase (Selkirk & Lee 2015). Much as how, under ROSE, spiking activity triggering clusters of linguistic features needs to be accessible to the O level via spike-phase coupling, structural syntactic information needs to be visible to systems of phonology and any associated probabilistic apparatus.

I will not stake a claim here about which precise pieces of syntactic information (e.g., phases) need to be accessible to phonology, only that there is at a minimum some interface (Nevins 2022; Newell & Piggott 2011). Some theorists hold that syntactic phases in the CP (Complementizer Phrase) and *v*P ('little' Verb Phrase) domains correspond to two phonological phrases, such that interpretive syntactic components are in effect aided in parsing by clear phonological events (Dobashy 2003; Ishihara 2007). Unlike syntactic tree structures, phonological representations consist of strings that lack generalized forms of recursion (Neeleman & Koot 2006). The major operations of phonology, like Copy, Align and Wrap (Scheer 2011), though operating within prosodic and syntactic phrase boundaries, do not require labeling.

Some of these mechanisms differ across languages (e.g., English exhibits focus-driven pitch accent placement; Japanese exhibits syntax-driven pitch reset at phrase boundaries). Frequency of use also impacts things like phonetic reduction ('the' is pronounced like 'thuh', not 'thee'). Probabilistic information affects prosodic boundary tones, pauses, and pitch resets, aligning them more closely with syntactically less predictable structures.

*3.2. Symbolic and Predictive Components: Neurobiological Factors*

Turning now to neurobiological factors, recall that phonological operations are highly sensitive to probabilistic information, and consider how phonetic information, surprisal and sequence statistics can be recovered from single-unit recordings (Leonard et al. 2024), indicating a possible architecture whereby coordinated spiking activity is managed by high-frequency γ phase delivering surprisal-based predictions for phonetic and phonological representations stored across cell clusters. The kind of population code for phoneme sequence representations reported in Gwilliams et al. (2022) is ripe for interfacing with neural codes for syntax-semantics operating primarily at the level of coordinated low-frequency phase codes for delivering syntactically-informed biases for predictive coding of incoming sounds. Concretely, the joint content-temporal coding for speech in Gwilliams et al. (2022) needs to be structured by higher-order lexical and phrasal information, given foundational assumptions at the syntax-phonology interface.

Under ROSE, this integration of syntactic instructions feeding phonetic parsing would take the form of low-frequency δ and θ phase coupling (originating within frontotemporal syntax regions) with the local high-frequency activity in auditory regions, in turn driving the population code for speech representations. Cross-frequency coupling provides a means to *integrate predictions* (Fiebelkorn et al. 2013; Grabot et al. 2019) across levels of linguistic information, with phase-based predictions being tested against amplitude-based sensorimotor transformations. Cross-frequency coupling (*n:m*), as implemented in ROSE, serves to *segregate* cortical computation into distinct but interfacing processing regimes (Mateos & Perez Velazquez 2024), in contrast to equal freqeuncy (*1:1*) connectivity that serves information integration. In Murphy (2024), high frequency phase-phase coupling was considered to be a means by which information of equivalent formats/types could be transmitted across cognitive interfaces, but incorporating these insights from Mateos and Perez Valzquez (2024) the emerging picture becomes much clearer: High γ power serves a fundamental binding role, as has long been hypothesized, but in combination with the cross-frequency coupling code for phrase structure (Murphy 2024; Weissbart & Martin 2024) this permits a connection between lower-order statistical inferences encoded in γ to be mapped to a symbolic phase-coupled lower frequency code. As reviewed extensively elsewhere (Murphy 2015b, 2018, 2020b, 2024), this accords with the consensus that lower frequencies help segregate faster rhythms (Northoff & Huang 2017). To put it another way, ROSE can readily permit the injection of statistical inferences given that it already calls upon a cortical computational architecture that achieves a careful balance and interplay between the brain's joint goal of ensuring *stability* (SE in ROSE: symbolic phase code responsible for representing basic phrase structure and configurations like {Subject, v* {Root, Object}}) in the face of the parallel need to incorporate *diversity* (RO in ROSE; statistical inferences) whilst processing multiple sensorimotor transformations at once.

A feasible neurobiological mechanism can be found in how ROSE can actively and flexibly optimize phasal and amplitude-based parameters to become more directly attuned to syntactic error detection and reanalysis processes (Mandal et al. 2020), which are rendered at the spiking level and communicated to the O and S levels via phase-amplitude coupling. This doubtless involves, and triggers, processes of neuromodulation for the precision-weighting of prediction errors (Möhring & Gläscher 2023), for example in cases of visual-orthographic prediction error representations (Fu & Gagl 2024). Through phase-amplitude coupling, ROSE also permits the modulation of 'synfire' chains (sequentially activated spiking patterns) for symbolic sequences (Zheng & Triesch 2014). Reflecting the ever-flexible nature of symbolic knowledge, we do not need to assume that frequency bands here are strict types with rigid functional interpretations; rather, they are more likely to be what Martin (2020: 1423) calls "tokens of processes with physiological bounds that render them into functional types".

I have previously discussed the potential role of ephaptic coupling (Murphy 2015b) in an emerging neural infrastructure for language, and with the introduction of ROSE we can now return to considering how this is concretely realized for reporting the statistical inferences over language-relevant stimuli. Ephaptic coupling, by influencing neural activity via extracellular fields (rather than via direct synaptic connections), provides a substrate for encoding probabilistic relationships, modulating signal-to-noise ratios, and shaping population-level dynamics. These effects align with statistical inference processes (Friston 2010). Groups of neurons may collectively encode stimuli features, with ephaptic coupling ensuring that the shared extracellular field conveys information about the combined activity of these neurons (Anastassiou

& Koch 2015; Pinotsis & Miller 2023). Properties of bioelectric fields could complement synaptic transmission and communication at the R and O levels of ROSE by representing language-relevant properties of statistical inference. Emerging bioelectric fields can sculpt neural activity and aid the propagation of information across cortical sites (Pinotsis & Miller 2023), making ephaptic coupling relevant to the mapping of lower-level statistical properties to global, network-level symbolic inferences. As Pinotsis and Miller (2023: 9878) put it, "the electric field enslaves neurons, not the other way around". A similar logic motivates ROSE, where lower frequencies act to scaffold local neural dynamics. Low-frequency oscillations create rhythmic fluctuations in the extracellular electric field, which can phase-align membrane potentials and synchronize local ephaptic coupling. Perhaps most obviously, low-frequency oscillations can modulate higher-frequency γ activity associated with ephaptic coupling, dynamically tuning its effects in a context-dependent manner. Oscillations can provide a temporal structure for updating predictions and errors (coded, for example, in Bayesian inference neural mechanisms; see Sohn & Narain 2021), with ephaptic coupling playing a role in local adjustments.

This helps us constrain further predictions from ROSE about single-unit responses. Direct cortical recordings could explore how representations from artificial language models pertaining to lexical statistics and dependency formation correlate with single-unit response profiles (Dijksterhuis et al. 2024; Lakretz et al. 2024), while more interactional network-level dynamics encoded via ROSE-type phase codes might correlate with punctuated moments of hierarchical syntactic node formation (Weissbart & Martin 2024). For example, spike-phase coupling is a strong candidate for implementing neural spiking precision to syntactically-driven predictions, with γ naturally being a stronger candidate for phase coordination in most cases than the LFP (Leonard et al. 2023). Syntactic prediction (Ferreira & Qiu 2021) inevitably generates surprisal-related and entropy-related measures for lexical feature selection, opening up questions about psycholinguistic factors, which we now turn to.

3.3. *Symbolic and Predictive Components: Psycholinguistic Factors*

In everyday cases, the use of statistical inferences may often suffice to extract a type of shallow parse (a 'good-enough' parse) for linguistic structure (Frances 2024). There is considerable individual variation in the extent to which comprehenders engage in syntactic prediction during online processing (Ferreira & Qiu 2021), influenced by age, literacy skill, and working memory capacity, with likely variation in how 'eager' the parser is (Chow & Chen 2020). ROSE therefore needs to incorporate elements of symbolic and probabilistic computations, flexibly adapting to how individuals access and exploit the resources of their 'I-language' (Chomsky 1986: 22). As Ferreira and Qiu (2021) note, every case of semantic (next-word) prediction logically involves a syntactic prediction, since all words belong to a grammatical class, whereas syntactic predictions do not always yield strong semantic anticipation. If a stranger walks up to you and says "Those…", you can readily predict a plural noun phrase, but not so much the semantic content. Syntactic prediction is therefore a more general phenomenon (Matchin et al. 2017).

At the same time, many predictions can be successful simply by virtue of lexico-semantic associations. In sentences like 'The day was windy so the girl went out to fly a __', the existing associations between 'windy', 'fly' and the anticipated 'kite' might be sufficient such that the comprehender may not need to rely on any additional linguistic information (Jackendoff & Audring 2020). It is only in cases of very sparse lexico-semantic saturation that logico-syntactic information would be saliently recruited to satisfy general predictive processing. Notice that in the 'kite' example semantic prediction mechanisms will also prime words that do *not* fit the correct syntactic category (e.g., 'glide', 'soar'), and so syntactic prediction will always be used to narrow the search space. Symbolic forms of information driven by syntactic inferences appear to be at least as robust, if not more so, than semantic prediction (van der Burght et al. 2021). One approach here that has proven useful in recent years is Tree-Adjoining Grammar (TAG), which incorporates elements of both syntactic structure and lexico-semantic predictions during parsing (Demberg et al. 2013; Kroch & Joshi 1985; Momma 2023).

## 4. A Hybrid Neural Code for Syntax

Many previous models of natural language syntax have been either wholly symbolic (Chomsky 2013, 2023; Chomsky et al. 2019, 2023) or wholly probabilistic (Kwiatkowski et al. 2012). Some recent models have tried to incorporate both of these (Marcus 2001; Martin 2020; Murphy et al. 2024a), and it is in this spirit that I have approached this topic here (see also Pessoa 2022).

In this section, my treatment of the neural code for compositional syntax will follow to some extent the schema of neurosymbolic approaches in artificial intelligence and cognitive science research like Marcus (2001), Garnelo and Shanahan (2019), Edelman (2008), Alers-Valentín et al. (2023), Garcez et al. (2015), Phillips (2020), Baggio and Martin (2020), Graves et al. (2016) and Smolensky et al. (2022), and it is vaguely in the spirit of the 'physical symbol system hypothesis' (Newell & Simon 1976; Nilsson 2007). As Minsky (1990) argued, "we must develop systems that combine the expressiveness and procedural versatility of symbolic systems with the fuzziness and adaptiveness of connectionist representations". Many connectionist approaches exploit extremely large amounts of data, multiple training iterations, and backpropagation for model learning, which are arguably not biologically plausible. In contrast, my approach is directly akin to Nilsson's (2007) concession: "I grant the need for non-symbolic processes in some intelligent systems, but I think they supplement rather than replace symbol systems".

The approach I will take in this section begins with the following premise: Assessing symbolic versus associationist models of syntax requires going beyond viewing syntax as a monolithic computational regime (as in Mahowald et al. 2024). Instead, we at least have to reckon with the fact that syntactic structures can deliver vertical, hierarchical representations that comply directly with symbolic models (Dehaene et al. 2022), but that horizontally encoded syntactic information (e.g., linear morphosyntactic relations and agreement dependencies) is expressed overtly and are often recoverable from statistical inferences in LLMs (Diego-Simón et al. 2024), even if such horizontal information is ultimately constrained by hierarchical syntax. The brain is sensitive to both hierarchically organized linguistic information and sequentially organized information (Burroughs et al. 2021; Zhao et al. 2024b). Due to the fact that structural and sequential properties of language very often correlate (Takahashi & Lidz 2008), with transitional probabilities being a decently useful cue for phrase structure (though, infamously, not always; Chomsky 1957; Slaats & Martin 2023), it has proven difficult to confidently isolate them in neurobiology. For instance, information entropy can facilitate lexical processing (Karimi et al. 2024), supporting the idea that statistical information serves to cue appropriate symbolic inferences (Hale 2003; Martin 2020;

Takahashi & Lidz 2008). Some recent work examining morpheme binding to syntactic positions has emphasized the importance of distributed vector representations whilst, simultaneously, highlighting "the role of structural information as the crucial determinant in the retrieval process" (Keshev et al. 2024).

But instead of ROSE simply being the architecture for symbolic knowledge, with some entirely separate neural code for connectionist knowledge, we can inject into appropriate levels of ROSE the necessary components of probabilistic reasoning, as shown in recent literature (Dekydtspotter et al. 2024; Shi 2024; Weissbart & Martin 2024; Yeaton 2024; Yu et al. 2024). For example, we can invoke the following:

(1) At the most basic level, cortical hierarchies transfer modality-specific sensorimotor representations to modality-general sites.

(2) Oscillatory mechanisms facilitate temporal integration of statistical patterns into a phase-coupled top-down symbolic code (Luo & Ding 2024).

(3) Distributed representations in a connectionist architecture for lexical learning are coordinated over time by frontotemporal traveling waves (Weissbart & Martin 2024). $\delta - \beta$ phase-amplitude coupling in posterior temporal cortex is jointly sensitive to lexical surprisal and syntactic complexity (tree depth); consider also the sensitivity of $\beta$ power to syntactic prediction (Murphy et al. 2022b; Zioga et al. 2023). $\theta - \gamma$ phase-amplitude coupling is jointly sensitive to syntactic node closures (in posterior temporal cortex) and word-level entropy (in inferior frontal cortex), pointing to a feasible interface between symbolic structural inferences being cued by lexical statistics (Weissbart & Martin 2024).

(4) Phrase structure predictions generated by higher-order language sites could link to prediction errors at distributed cortical sites, indexed by spike-timing codes and precision-weighting of errors, where lower levels of ROSE (RO) can partly encode sensory and error signals fed back to higher levels.

(5) Interactions between lateral prefrontal and posterior middle temporal cortical sites with various cortico-subcortical loops could achieve integration between symbolic and probabilistic knowledge (Murphy et al. 2022a).

(6) The statistical structure of episodic memories might interface with linguistically-generated oscillatory representations via a $\theta$ phase code (Shi 2024) (at the E level).

(7) Spectral information like phase and oscillatory cycles could offer a brain-intrinsic coordinate system by which the output of neurocognitive patterns can be interpreted (van Bree 2024).

(8) Top-down predictions about higher-order structure imposed by ROSE could be implemented via firing rate functions, non-linear postsynaptic responses, and modulations in neuronal connectivity (Friston et al. 2017), buttressed by the hypothesis that overlapping patterns of population activity encode statistics that summarize both estimates and uncertainties about latent variables (Pitkow & Angelaki 2017).

(9) Properties of synaptic computation (Aitchison et al. 2021; Soltani & Wang 2010) would here execute probabilistic inference for lexico-semantic feature spaces, a promising candidate for a biophysical instantiation of the Bayesian decision rule (e.g., more uncertainty leading to greater variability in postsynaptic potential size). This Bayesian coding can also be seen at the circuit (Narain et al. 2018) and population (Dubreuil et al. 2022; Ganguli & Simoncelli 2014) levels. An interesting topic concerns when and where 'modular' or 'transform' strategies for Bayesian inference in the brain are harnessed for syntactic and semantic prediction (Sohn & Narain 2021).

More generally, the core architectural component of ROSE, phase-amplitude coupling, can: (1) interface with frequency-specific statistical encoding; (2) ensure that local lexical statistics are aligned with global phrase-level predictions; (3) provide probabilistic representations via phase encoding; (4) encode statistical interactions (e.g., joint probabilities); (5) and represent statistical gradients (e.g., shifting probabilities or contextual relevance).

### 4.1. A Neurosymbolic Model: Morphosyntax

With an eye to injecting into ROSE properties of statistical inference pertinent to probabilistic neural architectures, we can model how frontotemporal traveling waves (the S and E components) might deliver symbolic representations into a neural regime encoding statistical properties of lexical items and morphemes, as required for many morphosyntactic processes. To do this, we can mathematically represent phase-amplitude coupling between low-frequency traveling waves and high-frequency spiking activity.

First, we can represent the traveling wave as a low-frequency oscillation (Zhang et al. 2018). Assume this wave propagates along a spatial axis (where $t$ = time; $x$ = spatial position; $f_L$ = frequency of the low-frequency wave; $k$ = wavenumber, representing spatial frequency; $\phi L$ = phase offset; $A_L$ = amplitude of the low-frequency oscillation):

$$L(t,x) = A_L \cos(2\pi f_L t - kx + \phi L)$$

Let the spiking activity be represented as a sum of high-frequency oscillations encoding statistical information about morphemes through frequency, amplitude, or rate codes (where $f_H^i$ = frequencies of high-frequency components; $A_H^i(x)$ = amplitudes of the high-frequency components at position $x$; $\phi_H^i$ = phases of the high-frequency components):

$$H(t,x) = \sum_{i=1}^{N} A_H^i(x) \cos(2\pi f_H^i t + \phi_H^i)$$

The statistical properties of morphemes might be encoded in the distribution of $A_H^i$ or $f_H^i$ (for the feasibility of this hypothesis, see Lakretz et al. 2024). Phase-amplitude coupling would then link the phase of $L(t,x)$ to the amplitude of $H(t,x)$:

$$A_H^i(x) = A_H^i(x, \phi_L(t,x))$$

To model this, we can assume $A_H^i(x, \phi_L)$ depends sinusoidally on the phase $\phi_L(t,x)$ of the low-frequency traveling wave (where $A_0^i(x)$ = baseline amplitude of high-frequency activity; $A_1^i(x)$ = modulation strength of high-frequency activity by the traveling wave, $\phi_P$):

$$A_H^i(x, \phi_L) = A_0^i(x) + A_1^i(x) \cos(\phi_L(t,x) - \phi_P)$$

In a combined model, the full spiking activity now integrates statistical inferences and top-down symbolic information via the traveling wave:

$$H_{mod}(t,x) = \sum_{i=1}^{N} [A_0^i(x) + A_1^i(x) \cos(\phi_L(t,x) - \phi_P)] \cos(2\pi f_H^i t + \phi_H^i)$$

The amplitude of high-freqeuncy spiking activity is here modulated ('mod') by the phase of the traveling wave, achieving a minimal scheme for how a ROSE architecture can begin to integrate symbolic and statistical inferences. Next, we can assume that the statistical properties of morphemes are encoded in $f_H^i$ (e.g., lexical frequency or predictability) and $A_0^i(x)$ (e.g., baseline relevance). The traveling wave modulates these properties dynamically via phase-dependent changes in $A_1^i(x)$. For instance, $\phi_P$ might represent a phase where certain lexical features are prioritized (Murphy 2020b, 2024). $A_1^i(x)$ could vary with context in order to emphasize or suppress specific features of lexical items.

In this model, to inject symbolic information low-frequency spatiotemporal properties could encode structured, rule-based information (Murphy 2015b, Kazanina & Tavano 2023b). Symbolic features (e.g., syntactic labels, heads) can be represented by the spatial gradient ($kx$) of the wave, such that different cortical areas are modulated differently. With respect to phase timing ($\phi_L$), phase relationships between the traveling wave and high-frequency activity could define specific roles, such as predictive vs. error signals in predictive coding (Friston 2010).

This demonstrates how the phase code in ROSE provides global organization, akin to symbolic systems that impose rules over distributed data. Spiking activity encodes statistical properties of lexical features at the R level, such as lexical frequency (spiking rate modulation), contextual predictability (amplitude variations reflecting prior probabilities or surprise), and lexical associations (patterns of co-activation among spiking populations). The statistical nature of spiking activity aligns clearly with connectionist principles, being distributed, probabilistic, context-sensitive, emerging from local learning rules (e.g., Hebbian or gradient-based processes), and encoding features continuously and adaptively, rather than discretely.

In sum, cross-frequency coupling allows symbolic rules to flexibly shape connectionist representations, creating a contextualized hybrid model. This mirrors how symbolic AI systems impose rules on a neural network's learned patterns (Marcus 2001). Under ROSE and its various postulated interfaces with probabilistic mechanisms, symbolic representations and rules are implicitly embedded in phase dynamics. Importantly, under this model symbolic rules are imposed dynamically rather than being stored redundantly in the brain, allowing for compact (and *compressible*; Dehaene et al., 2022; Murphy et al. 2024a) representations.

## 4.2. A Neurosymbolic Model: Syntax-Semantics

Having demonstrated the biological feasibility of interfacing syntactic phase codes with lexical statistics, we can now turn to the interface of syntax and lexico-semantic properties. The mathematical model in this section will aim to demonstrate how population spiking activity encoding statistical information about semantics can interface with a low-frequency phase shift representing syntactic categories through phase-amplitude coupling, as in ROSE.

We can leverage here phase-of-firing codes, population rate dynamics, and phase-amplitude coupling to construct a phase-modulated population spiking model. First, let the low-frequency oscillation encode symbolic information about the syntactic category ($C$) of the phrase (e.g., Noun Phrase or Verb Phrase) (Kazanina & Tavano 2023a, 2023b). This oscillation is given as (where $f_s$ = frequency of the oscillation; $\phi c$ = phase shift that encodes the syntactic category ($C$); $A_s$ = amplitude of the low-frequency wave):

$$S(t) = A_s \cos(2\pi f_s t + \phi_C)$$

Symbolic information is here mapped to the phase $\phi_C$, distinguishing syntactic categories/labels.

Next, assume that a population of $N$ neurons encodes lexico-semantic statistics (Goldstein et al. 2024), ($P(S)$) of a linguistic phrase, through spiking activity. The spike train of the $i$-th neuron can be modeled as an inhomogeneous Poisson process (Chiu et al. 2013) (where $r_i(t)$ = spike rate of the $i$-th neuron; $\lambda_i(t)$ = baseline firing rate that reflects semantic statistical properties; $\eta_i(t)$ = noise term, e.g., Gaussian or white noise):

$$r_i(t) = \lambda_i(t) + \eta_i(t)$$

Relevant lexico-semantic statistics can be distributed across the population (where $P(S_i)$ represents the semantic weight of a specific feature):

$$\lambda_i(t) = P(S_i) \quad \text{for } i = 1, \dots, N$$

As with the model above (sub-section 4.1), we can then invoke phase-amplitude coupling as an interface between symbolic and statistical/lexico-semantic representations. Let the spike rate of each neuron $r_i(t)$ be modulated by the phase of the low-frequency oscillations $S(t)$ (where $\alpha$ = coupling strength between spiking and low-frequency phase which could be learned through a Hebbian rule; $\phi_C(t)$ = instantaneous phase of the low-frequency oscillation; $\phi_P$ = preferred phase of firing for the population):

$$\lambda_i(t) = P(S_i)\,[1 + \alpha \cos(\phi_C(t) - \phi_P)]$$

This equation ensures that the spike rates are amplified/suppressed depending on the syntactic category being encoded in $\phi_C$.

We can now model the total spiking activity across the population, reflecting the interface between lexico-semantic statistics at R and O levels and the syntactic category at S and E levels:

$$R(t) = \sum_{i=1}^{N} \lambda_i(t) = \sum_{i=1}^{N} P(S_i)\,[1 + \alpha \cos(\phi_C(t) - \phi_P)]$$

This summation aggregates the lexico-semantic statistics, modulated by the syntactic category encoded in the ROSE phase code.

Both the models presented here, and above (4.1), emphasize slightly different dynamics and means of integrating statistical information with symbolic representations that accord with ROSE and empirically documented mechanisms in the neurolinguistics literature. Symbolic systems here are the ones that actively interface with connectionist parsing mechanisms – and not the other way round.

### 4.3. An Integrated Model for ROSE

We can tentatively begin to map out parts of an integrated mathematical model for ROSE by modeling R as compressed/reduced lexical feature representations (Murphy et al. 2024a), O as population dynamics representing the binding of these lexical features, S as a tree-structured oscillatory hierarchy, and E as phase-aligned traveling waves.

Alongside the relevant models for R provided above, we can also consider the need to eliminate redundant information to simplify basic lexical item representations (Dehaene et al. 2022). Let $X \in \mathbb{R}^n$ represent a high-dimensional vector encoding lexical features (Murphy 2024), such as word embeddings or feature bundles. We can apply a transformation, $R(X)$, to remove redundant dimensions (where $W \in \mathbb{R}^{m \times n}$ = projection matrix ($m < n$)):

$$R(X) = WX$$

Reduction could occur in a recurrent network (where $h_t$ represents a compressed representation at time *t*):

$$h_t = \sigma(W_r h_{t-1} + W_x X_t + b)$$

This is effectively a way to ground the increasingly plausible notion that the barcode or algorithmic vector symbolic architecture (Eliasmith et al. 2012; Murphy 2024) for lexico-semantic features being accessed by S complies with demands of minimal description (Dehaene et al. 2022; Piantadosi & Gallistel 2024).

These lexical feature bundles at R can be represented at O ($O_k$) by γ-coordinated (spike-phase coupled) synchronously firing neural populations (where $r_i(t)$ = spike train of neuron $i$; $O_k(t)$ = firing rate of the population encoding the $k$-th syntactic object accessible to the S and E levels):

$$O_k(t) = \sum_{i \in \text{Population}_k} r_i(t)$$

We can model S using traditional tree structures (Murphy et al. 2024a) or graph embeddings. For instance, we can represent the syntax of a phrase as a tree $T$, where nodes correspond to syntactic objects and edges encode relationships (e.g., dominance, dependency), as in standard generative grammar (where $V$ = set of vertices; $E$ = set of edges):

$$T = (V, E)$$

We can also use recursive neural networks (Chowdhury & Caragea 2021) to encode hierarchical structures at S – or, for bonus psycholinguistic plausibility, left-corner recurrent neural network grammars (Sugimoto et al. 2024, but see Slaats et al. 2024; I suspect that different parsing models will have their performance impacted by whether we apply them to low-frequency components or high-frequency $\gamma$; e.g., bottom-up parsing for $\gamma$ but left-corner for slower oscillations) (where $f$ is a nonlinear function):

$$h_{\text{parent}} = f(h_{\text{left\_child}}, h_{\text{right\_child}})$$

ROSE assumes that multiple types of nestings yield recursion at S (Murphy 2024, Fig. 1), which can be modeled as successive operations on oscillatory phases, where each iteration creates a new syntactic structure (where $\Delta\phi$ = the addition of a new constituent, i.e., via a new $\theta$ complex into the $\delta$ phase through recurrent phase-amplitude coupling interactions between frontotemporal brain regions):

$$\phi_{nested}(t) = \phi_{matrix}(t) + \Delta\phi$$

Lastly, adding to what we have already established in the above models for phase-amplitude coupling, as discussed in Murphy (2020b) phase coupling at the S and E levels can be described using Kuramoto models for coupled oscillators (where $\phi_i$ = phase of oscillator $i$; $K_{ij}$ = coupling strength between oscillators $i$ and $j$):

$$\frac{d\phi_i}{dt} = \omega_i + \sum_j K_{ij} \sin(\phi_j - \phi_i)$$

The state representation of ROSE at time $t$ would evolve under coupled differential equations:

$$\frac{dR}{dt} = F(R, t)$$

Where:

$$F(R, t) = \begin{bmatrix} -\nabla WX \\ \sum_{i \in \text{Populations}} r_i(t) \\ f_{\text{tree}}(h_{\text{parent}}, h_{\text{children}}) \\ \sum_j K_{ij} \sin(\phi_j - \phi_i) \end{bmatrix}$$

The four components of ROSE are here stacked top to bottom.

This mathematical model grounds key elements of ROSE in formal neural dynamics, by combining dimensionality reduction, high-frequency population spiking, hierarchical symbolic structures, and low-frequency phase codes, yielding a biologically plausible foundation for the representation of compositional syntax as a hybrid of symbolic and connectionist neural processes.

### 4.4. Modeling MERGE via Hopf Algebra

Next, I will take as a case study Marcolli et al. (2025) and their model of MERGE via Hopf algebra, which provides a useful framework for representing hierarchical and recursive processes. Marcolli et al. (2025) prove that MERGE-based derivations are strictly Markovian, rendering a potential mapping to neural codes feasible. Likewise, we can model via ROSE this particular MERGE-based system. Hopf algebra involves multiplication ($M$), comultiplication ($\Delta$), amongst other things, but I will here keep to the essentials for modeling some features of phase structure generation.

An oscillatory component can be represented as an element $x$ (where $A$ = amplitude; $\phi$ = phase):

$$x = A_\gamma e^{i\phi\gamma}$$

The operation MERGE, $M(x, y)$, combines oscillatory states, forming a hierarchical structure:

$$M(x, y) = A_\gamma A_\theta e^{i(\phi\gamma + \phi\theta)}$$

Comultiplication would decompose an oscillatory state into its components (where $\delta$ phase-modulates $\theta$ amplitude, and $\theta$ phase-modulates $\gamma$ amplitude) (Murphy 2024):

$$\Delta(x) = x \oplus z = (A_\gamma e^{i\phi\gamma}) \oplus (A_\delta e^{i\phi\delta})$$

Hierarchical recursion is modeled here when the product of two oscillatory states feeds back into the system:

$$x_{n+1} = M(x_n, z) = (A_\gamma^n A_\delta) e^{i(\phi_\gamma^n + \phi_\delta)}$$

There are other potential sympathies between the MERGE-based syntax in Marcolli et al. (2025) and ROSE, which I will leave for future investigation.

### 4.5. The Unbearable Lightness of Language Models

In this search for a hybrid architecture for syntax, another point of possible connection lies in transformer architectures and their positional embeddings (Li et al. 2023). As widely discussed, these provide evolving projections that allow the model to locate each token in a sequence. The general dynamics of this process align with the need to track the evolving, embedded representations in a sentence such as during the tracking of filler-gap dependencies and other forms of horizontally encoded syntax, whilst preserving the order and resolution of existing information at lower linguistic levels (Gwilliams et al. 2024).

While deep language models succeed at what I am framing as horizontal components of syntactic information (Linzen & Baroni 2021), they seem to perform much more poorly at capturing deep syntactic structures encoding vertical phrase structure and compositional properties (Dentella et al. 2024; Lakretz et al. 2021; Manning et al. 2020). Even though long-distance dependencies, like in English subject-verb agreement, are encoded via structural relations and not linear relations, the distributional properties of the more common forms in English appear sufficient for LSTMs to represent them fairly well (Gulordava et al. 2018). This is not the case for infrequent sentence types, such as nested agreement dependencies across an object relative clause (Marvin & Linzen 2018), which shifts the burden back to invoking vertical processing biases.

Others have argued that what appears to be an effect of syntactic islands (a clear example of vertical structure imposing constraints on parsing) on language model probabilities can instead be explained using nongrammatical factors (Chowdhury & Zamparelli 2018). Deep neural networks do not appear to capture the full complexity of island constraints, a classic example of vertically encoded syntax; e.g., they do not capture negative islands where syntactic instructions to pragmatics play a role ('*How fast didn't John drive__?') (Chaves 2020). Researchers that have documented promising results for deep neural networks with respect to various aspects of syntax do not find this level of sensitivity for island effects (Warstadt et al.

2020). Using the seq2seq framework, McCoy et al. (2020) showed a clear linear-based bias for English auxiliary fronting, in contrast to human structure-dependence of rules.

Meanwhile, word surprisal can successfully predict the existence of garden-path effects during human self-paced reading – but it drastically underpredicts the magnitude of these effects, and it also fails to predict their relative severity across constructions (van Schijndel & Linzen 2021). This suggests that a syntactic repair or reanalysis mechanism is needed, not just information-theoretic inferences.

A more general conclusion we can make here is that linguistic meaning composition *does not seem to be adequately accounted for exclusively by statistical relationships or associative processing* (Baggio 2018) – in stark contrast to many other cognitive systems and perception-action cycles.

In summary, current connectionist models of syntax typically exhibit a strong bias for syntactically simple linear-distance units (Lakretz et al. 2019). This might account for why they perform well on horizontal linguistic representations but less well on vertical, hierarchical structure – buttressed by their facility for sequential left-to-right processing and content-addressable memory storage and retrieval (gating, attention). A recent test of sample-efficient pretraining on developmentally plausible amounts of data (via the BabyLM Challenge) does not inspire confidence concerning robust vertical syntactic representations (Warstadt et al. 2023). Another major issue is the very limited ability for *generalization* with deep neural networks (Pitkow 2023; Qi et al. 2024; Zhang et al. 2024), which is a hallmark trait of applying vertical symbolic/categorial rules to novel distributions. Even Apple AI found no evidence of formal reasoning in language models, but rather sophisticated (and fragile) pattern matching (Mirzadeh et al. 2024).

It is my contention, then, that the neural code for syntax will be neurosymbolic, capturing the layered nature of processing, where lower layers process distributed representations and higher layers handle abstract, symbolic information. This requires the integration of two paradigms: symbolic (rule-based, discrete) for hierarchical vertical syntax, and connectionist (distributed, gradient-based) for syntax-phonology relations and linear morphosyntactic relations and dependencies.

Relatedly, the framework outlined here helps re-focus the use of LLMs in cognitive neuroscience research. Instead of being deployed universally across all types of linguistic processes (perhaps as part of an effort to "refute" modern linguistic theory; Piantadosi 2024), the present framework assumes that LLMs will be highly productive resources for exploring aspects of the syntax-phonology interface, or the externalization and linearization of syntactic dependencies and forms of morphosyntactic prediction. For instance, they can be used successfully to correlate brain embeddings with artificial contextual embeddings pertaining to lexical features (Goldstein et al. 2024). In contrast, forms of vertical syntactic information like labeled phrase structures and compositional reasoning involving headedness (Collins 2019; Pietroski 2018) will be grounded in low-frequency phase codes that flexibly adapt their tuning to lower-order sensorimotor features via spike-phase coupling.

I refer the reader to Dentella et al. (2024) for a comprehensive assessment of the grammatical and semantic competence of LLMs, but here I will simply note my position that the fluency of LLM output and their simultaneous failure with basic compositional reasoning suggest that horizontal dimensions of syntactic relations may be encoded, but even the simplest form of minimal phrasal composition may be absent or poorly represented, given "little evidence of sophisticated phrase composition" (Yu & Ettinger 2020) in artificial models. Studies of human reading that explicitly probe for syntactic disambiguation effects show that probabilistic measures like surprisal are not sufficient to explain task difficulty (Huang et al. 2024).

That the brain uses probabilistic information to infer syntactic structure is not altogether surprising, but it is of considerable interest that these information sources can be resolved together to aid parsing via low-frequency δ activity (Slaats et al. 2024), and that bottom-up node counts (as opposed to top-down or left-corner models) received the greatest empirical support in Slaats et al. (2024) in terms of contributing to modeling the δ signal. This indicates the possibility for some (yet to be determined) low-frequency phase code for generating statistical and symbolic predictions (perhaps configurational predictions for syntactic structure).

Zhao et al. (2024b) rightly "call for a neurobiological account of language processing wherein the brain leverages regularities computed over multiple levels of linguistic representation to guide rhythmic computation, and which seeks to integrate the contributions of structural and sequential information to language processing, and to behavior more generally". It is my hope that this Discussion article provides a direction for how this can be achieved: a core symbolic architecture (ROSE) can be seen to be sensitive to the statistics of natural language via mechanisms modeled above. I leave it to future research to uncover which precise instantiation of ROSE-compliant interfaces with statistical knowledge is most empirically supported and conceptually parsimonious. For instance, I have focused here on mechanisms in physical space (traveling waves, population spiking), but models that incorporate state space (Glaser et al. 2020) and metastable network dynamics (Forseth et al. 2021) will likely also be needed to fully articulate the interactions between statistical inferences and symbolic state transitions.

## 5. Conclusion

> Happy is he who can with a vigorous wing
> Propel towards the luminous and serene realms;
> He whose thoughts, like larks,
> Free, in the morning take flight,
> — Hover over life, and understand with ease
> The language of flowers and silent things!
> Charles Baudelaire, 'Élévation', *Les Fleurs du mal* (1861)

A number of avenues – experimental and theoretical – have been provided to build a hybrid neurosymbolic model of syntax in the brain, anchored around the ROSE neurocomputational architecture. When discussing symbolic models versus neural nets, Dehaene et al. (2022) note that "[a] middle ground must be found" (as argued also in Marcus & Murphy 2022). It is my intention that the mathematical models provided here demonstrate a plausible interplay between a low-frequency phase code and statistically-driven neural mechanisms that can offer a framework where symbolic and connectionist representations can interface dynamically, leveraging the strengths of both paradigms. Nevertheless, this proposal is a model, like any other, with clear limitations in terms of its expressive scope for natural linguistic behavior, focusing narrowly only on the immediate integration of basic hierarchical phrase structure with directly connected features related to the statistics of relevant lexical items and their dependencies.

I am not the first to notice that the cognitive neuroscience of language has recently been populated by a mixture of technically generic but rhetorically majestic proposals that lack much in the way of concrete neurobiological grounding. We read about *dynamic* syntactic parsing, the *neural geometry* of semantics, *deep language modelling* of lexico-semantics, *neural readouts* of grammar – much of which is injected into experimental results rather than emerging more casually from a formal framework. A reasonable complaint from Piantadosi (2021) is that while "[t]here are many connectionist approaches that try to explain—or explain away—symbolic thought", in contrast "*almost no* work on the symbolic side has sought to push down towards representations that could more directly be implemented". The hybrid model I have developed here constitutes such an example of symbolic theory aiming to make room for connectionist processes. Rather than remaining "drunk on symbols" (as Richard Dawkins recently put it to Jordan Peterson), a hybrid neural architecture that interfaces symbolic knowledge with statistical learning will be essential if the field is to move beyond some of its less productive and contentious debates. The recent Nobel Prize in Chemistry to the AlphaFold team represents a major success for neurosymbolic approaches (Callway 2024). Hybrid models of syntax in the brain can help unify perspectives that rightly emphasize the existence of linguistic prediction (Forseth et al. 2020) with frameworks that highlight the role of language in inference generation, cognitive/generative model updating, consolidation of experience, endogenous planning and monitoring, aiding directed attention, reflection of personal experience and values, and compositional semantics.

Developing hybrid models will enhance our ability to more confidently isolate neural mechanisms of syntax that would otherwise be invisible to direct explication, and that are de facto shielded in the current research climate by the growing reach of deep learning that continues to cloak the neural code for language in shadow.

## Acknowledgments

My thanks go to Meredith J. McCarty and Mitchell B. Slapik for helpful comments on an earlier draft, and to Sander van Bree, Jake Quilty-Dunn and Evelina Leivada for useful discussion.